%% file: main.tex
\documentclass[letterpaper, 10 pt, conference]{ieeeconf}
\IEEEoverridecommandlockouts
\usepackage{amsmath}
\usepackage{amssymb}             
\usepackage{amsfonts}            
\usepackage{mathrsfs}            

\usepackage{amsthm}
\usepackage{graphicx}
\usepackage{float}
\usepackage{booktabs}
\usepackage{multirow}
\usepackage{outline}
\usepackage{paralist}
\usepackage{siunitx}
\usepackage[table, dvipsnames, xcdraw]{xcolor}
\usepackage{caption}
\usepackage{adjustbox}
\usepackage{xspace}
\usepackage{cite}
\usepackage{balance}

\usepackage{hyperref}

\input{math_commands}

\newcommand{\ours}[0]{LCP\xspace}

\theoremstyle{definition}
\newtheorem{definition}{Definition}[section]

\definecolor{deepblue}{HTML}{27a2c3}

\title{\LARGE{\textbf{Learning Smooth Humanoid Locomotion\\ through Lipschitz-Constrained Policies}}
}

\author{
Zixuan Chen$^{* 1}$\quad Xialin He$^{* 2}$\quad Yen-Jen Wang$^{* 3}$\quad Qiayuan Liao$^{3}$\quad Yanjie Ze$^{4}$\quad Zhongyu Li$^{3}$\quad \\ S. Shankar Sastry$^{3}$\quad Jiajun Wu$^{4}$\quad Koushil Sreenath$^{3}$ \quad Saurabh Gupta$^{2}$\quad Xue Bin Peng$^{1,5}$\vspace{0.03in}\\
$^1$Simon Fraser University\quad $^2$UIUC \quad $^3$UC Berkeley\quad $^4$Stanford University\quad $^5$NVIDIA \quad $^*$Equal Contribution\vspace{0.03in}\\
\href{https://lipschitz-constrained-policy.github.io}{\color{deepblue}\textbf{lipschitz-constrained-policy.github.io}\xspace}\vspace{-0.1in}
}

\begin{document}

\twocolumn[{%
\renewcommand\twocolumn[1][]{#1}%
\maketitle
\begin{center}
    \centering
    \captionsetup{type=figure}
    \includegraphics[width=0.98\textwidth]{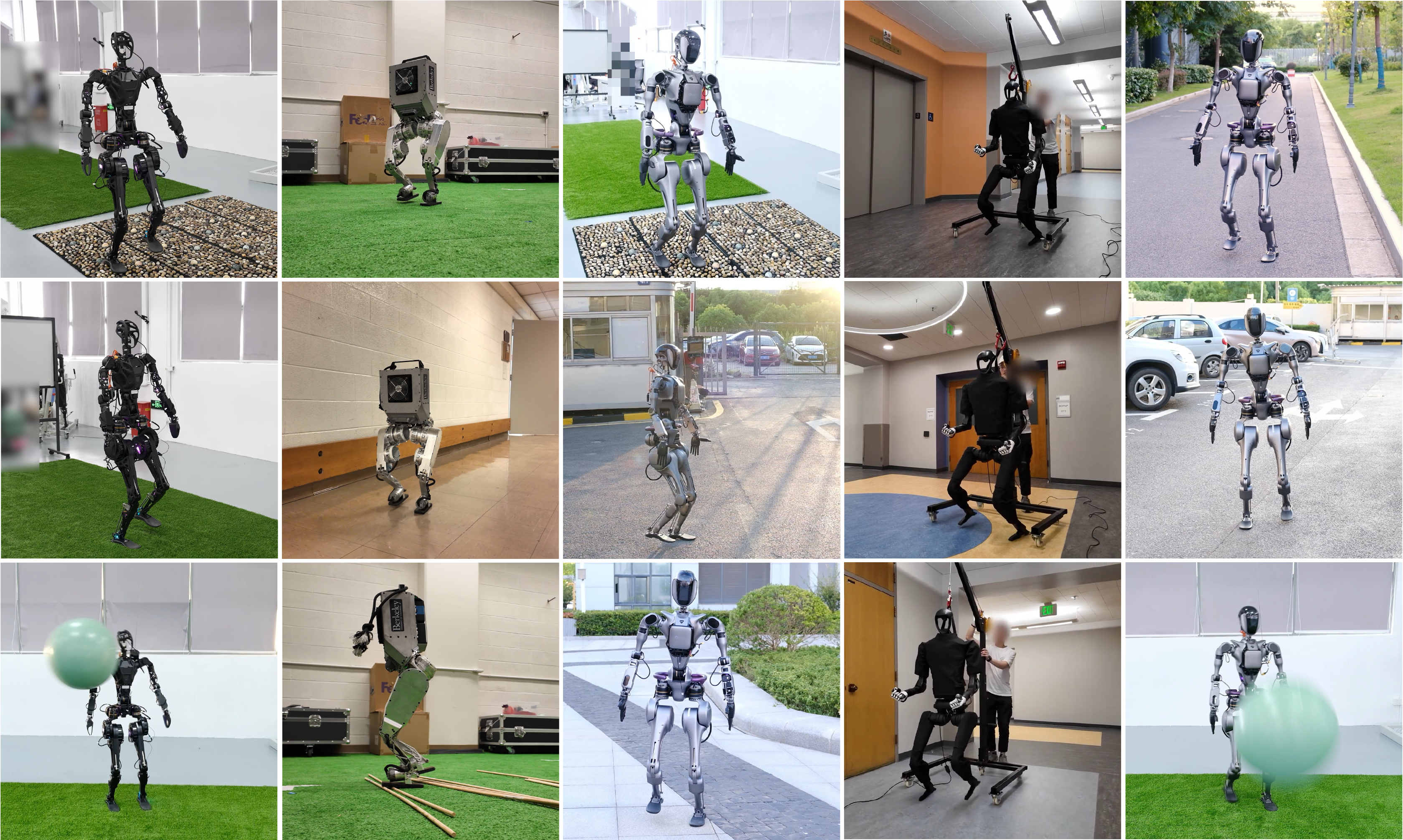}
    \caption{Lipschitz-constrained policies (\ours) provide a simple and general method for training policies to produce smooth behaviors, which can be directly deployed on a wide range of real-world humanoid robots. Our policies exhibit robust behaviors that can recover from external forces and walk across irregular terrain. For full videos, please visit the \href{https://lipschitz-constrained-policy.github.io}{project website}.}
    \label{fig:teaser}
\end{center}
}]

\begin{abstract}
Reinforcement learning combined with sim-to-real transfer offers a general framework for developing locomotion controllers for legged robots. To facilitate successful deployment in the real world, smoothing techniques, such as low-pass filters and smoothness rewards, are often employed to develop policies with smooth behaviors. However, because these techniques are non-differentiable and usually require tedious tuning of a large set of hyperparameters, they tend to require extensive manual tuning for each robotic platform.
To address this challenge and establish a general technique for enforcing smooth behaviors, we propose a simple and effective method that imposes a Lipschitz constraint on a learned policy, which we refer to as Lipschitz-Constrained Policies (\ours). We show that the Lipschitz constraint can be implemented in the form of a gradient penalty, which provides a differentiable objective that can be easily incorporated with automatic differentiation frameworks. We demonstrate that \ours effectively replaces the need for smoothing rewards or low-pass filters and can be easily integrated into training frameworks for many distinct humanoid robots. We extensively evaluate \ours in both simulation and real-world humanoid robots, producing smooth and robust locomotion controllers. All simulation and deployment code, along with complete checkpoints, is available on our project page: \href{https://lipschitz-constrained-policy.github.io}{\color{deepblue}\textbf{https://lipschitz-constrained-policy.github.io}\xspace}.
\end{abstract}


\section{Introduction}

Humanoid research aims to develop intelligent, human-like machines capable of autonomously operating in everyday environments \cite{cheng2024open,ze2024humanoid_manipulation, radosavovic2024learninghumanoidlocomotionchallenging, liu2024opt2skill}. One of the most fundamental challenges in this field is achieving reliable mobility. Developing robust locomotion controllers and adapting them to real robots would greatly improve their capabilities.

Traditional model-based methods, such as Model Predictive Control (MPC), necessitate precise system structure and dynamics modeling, which is labor-intensive and challenging to design. In contrast, model-free reinforcement learning provides a straightforward end-to-end approach to developing robust controllers, significantly alleviating the necessity for meticulous dynamics modeling and system design. However, because model-free RL requires a large number of samples through trial-and-error during training, which cannot be performed in the real world, sim-to-real transfer techniques are utilized to enable the successful deployment of controllers in real-world environments.
Combined with sim-to-real techniques, model-free RL-based methods have achieved great success in controlling quadruped robots and humanoid robots \cite{rudin2022learning, fu2023deep, cheng2024expressive}.

However, due to the simplified dynamics and actuation models used in simulation, the resulting models tend to be nearly idealized, meaning that the motors can produce the desired torques at any state. 
As a result, RL-based policies trained in simulation are susceptible to developing jittery behaviors akin to bang-bang control \cite{BangBangLASALLE1960}. This results in significant differences between actions in consecutive timesteps, leading to excessively high output torques that real actuators cannot produce. As such, these behaviors often fail to transfer to real robots. Therefore, enforcing smooth behaviors is crucial for successful sim-to-real transfer.

Previous systems use smoothing methods to enforce smooth behaviors from a learned policy, such as smoothness rewards or low-pass filters. Incorporating smoothness rewards during training can be an effective approach to eliciting smoother behaviors. In robot locomotion, researchers typically penalize joint velocities, joint accelerations, and energy consumption \cite{fu2021minimizing}. Other approaches attempt to smooth policy behavior by applying low-pass filters \cite{ji2022hierarchical, li2021reinforcement}. However, smoothness rewards require careful tuning of weights to balance smooth behavior with task completion, and low-pass filters often dampen or limit exploration, causing extra effort when training controllers for a new robot. Additionally, the non-differentiable nature of smoothness rewards and low-pass filters presents another limitation.

In this work, we introduce Lipschitz-Constrained Policies (\ours), a general and differentiable method for encouraging RL policies to develop smooth behaviors. 
\ours enforces a Lipschitz constraint on the output actions of a policy with respect to the input observations through a differentiable gradient penalty. \ours can be implemented with only a few lines of code and easily incorporated into existing RL frameworks. We demonstrate that this approach can be directly applied to train control policies for a diverse suite of humanoid robots. Our experiments show that \ours can be an alternative to non-differentiable smoothness techniques such as smoothness rewards and low-pass filters. We also demonstrate that \ours can be deployed zero-shot to several real-world robots with different morphologies, indicating the generalization of our method.




\section{Related Work}

Legged robot locomotion has long been a crucial yet challenging problem in robotics due to legged systems' high dimensionality and instability. Classic model-based control methods have achieved impressive behaviors on legged robots \cite{miura1984dynamic, sreenath2011compliant, geyer2003positive}. 
In recent years, learning-based methods have shown great potential to automate the controller development process, providing a general approach to building robust controllers for quadrupedal locomotion \cite{cheng2023parkour, kumar2021rma, lai2023sim, peng2020learning}, bipedal locomotion \cite{li2021reinforcement,li2023robust,kumar2022adapting,duan2023learning}, and humanoid locomotion \cite{radosavovic2024real, gu2024advancing, cheng2024expressive, he2024omnih2o}.

\paragraph{Sim-to-Real Transfer}
One of the main challenges in RL-based methods is sim-to-real transfer, where policies are first trained in simulation and then deployed in real-world environments. Substantial effort is often necessary to bridge the domain gap between simulations and the real world, such as developing high-fidelity simulators \cite{todorov2012mujoco, makoviychuk2021isaac}, and incorporating domain randomization techniques during training \cite{Sim2Real2018, peng2020learning, radosavovic2024humanoid, radosavovic2024real}. Another widely adopted approach is the teacher-student framework, where a privileged teacher policy, with access to full state information, is trained first, followed by the training of an observation-based student policy through distillation \cite{gu2024humanoid, cheng2023parkour, zhuang2024humanoid, he2024omnih2o, radosavovic2024real, kumar2022adapting, fu2024humanplus}. To further facilitate sim-to-real transfer, our framework also leverages a teacher-student framework \cite{kumar2021rma, fu2023deep, gu2024advancing}, which trains a latent representation of the dynamics based on the observation history. These methods have been successful in transferring controllers for both quadruped robots \cite{liu2024visual, cheng2023legs, fu2021minimizing}, and humanoid robots \cite{cheng2024expressive, gu2024advancing}. Some work also explores utilizing a single policy to control robots with different morphologies zero-shot in real world \cite{bohlinger2024one}. However, the policy's performance on real humanoid robots has yet to be validated, and it is not easy to plug into any existing training pipeline. 

\paragraph{Learning Smooth Behaviors}
Due to the simplified dynamics of simulators, policies trained in simulation often exhibit jittery behaviors that cannot be transferred to the real world. Therefore, smooth policy behaviors are critical for successful sim-to-real transfer. Common smoothing techniques include the use of smoothness rewards, such as penalizing sudden changes in actions, degree of freedom (DoF) velocities, DoF accelerations \cite{liu2024visual, fu2024humanplus, he2024omnih2o, he2024learning, gu2024humanoid, zhang2024wococo}, and energy consumption \cite{fu2023deep, fu2021minimizing}. In addition to smoothness rewards, low-pass filters have also been applied to the output actions of a policy to ensure smoother behaviors \cite{peng2020learning, li2021reinforcement, ji2022hierarchical, feng2023genloco}. However, smoothness rewards typically require careful manual design and tuning, while low-pass filters often dampen policy exploration, resulting in sub-optimal policies. These techniques are also generally not directly differentiable, requiring sample-based gradient estimators to optimize, such policy gradients.

\paragraph{Gradient Penalty}
In this work, we propose a simple and differentiable method to train RL policies that produce smooth behaviors by leveraging a gradient penalty. Gradient penalty is a common technique for stabilizing training of generative adversarial network (GAN), which is susceptible to vanishing or exploding gradients. \textit{Arjovsky et al.} \cite{arjovsky2017wassersteingan} proposed the Wasserstein GAN (WGAN) using weight clipping to stabilize training. However, weight clipping still often results in poor model performance and convergence issues. \textit{Gulrajani et al.} \cite{gulrajani2017improvedtrainingwassersteingans} introduced the gradient penalty (WGAN-GP) as an alternative to weight clipping, which penalizes the norm of the discriminator’s gradient. Since its introduction, the gradient penalty has become a widely used regularization technique for GANs \cite{karras2018progressivegrowinggansimproved,brock2019largescalegantraining}.
For motion control, gradient penalty has been an effective technique for improving the stability of adversarial imitation learning. For example, AMP \cite{Peng_2021}, CALM \cite{tessler2023calm}, and ASE \cite{Peng_2022} all apply a gradient penalty to regularize an adversarial discriminator, which then enables a policy to imitate a large variety of challenging motions. While these prior systems demonstrated the effectiveness of gradient penalties as a regularizer for discriminators, in this work, we show that a similar gradient penalty can also be an effective regularizer to encourage policies to produce smooth behaviors, which are then more amenable for real-world transfer.


\section{Background}
Our method leverages ideas from Lipschitz continuity to train reinforcement learning policies to produce smooth behaviors. This section will review some fundamental concepts for Lipschitz continuity and reinforcement learning to provide a comprehensive background of our proposed method.

\subsection{Lipschitz Continuity}
\begin{figure}[t]
    \centering
    \includegraphics[width=0.98\columnwidth]{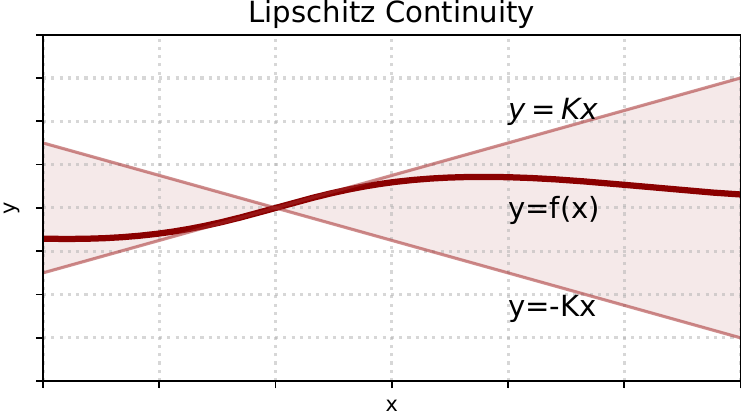}
    \caption{Lipschitz continuity is a method of quantifying the smoothness functions. A Lipschitz continuous function is a function whose rate-of-change is bounded by a constant $K$.}
    \label{fig:vis-lipschitz-cont}
\vspace{-0.25cm}
\end{figure}
Intuitively, Lipschitz continuity is a property that limits how fast a function can change. This property is a good way of characterizing the smoothness of a function. An intuitive visualization is shown in \Figref{fig:vis-lipschitz-cont}. Formally, we give the definition of Lipschitz continuity as follows:
\begin{definition}[Lipschitz Continuity]
    Given two metric spaces $(X, d_{X})$ and $(Y, d_{Y})$, where $d_{X}$  denotes the metric on the set $X$ and $d_{Y}$ is the metric on set $Y$, a function $f: X\rightarrow Y$ is deemed \textbf{Lipschitz continuous} if there exists a real constant $K$ such that, for all $\rvx_1$ and $\rvx_2$ in $X$,
   \begin{equation}
     d_{Y}(f(\rvx_1), f(\rvx_2)) \leq K d_{X}(\rvx_{1}, \rvx_{2}).
    \label{eq:def-lipschitz}
    \end{equation}
    Any such $K$ is referred to as a \textbf{Lipschitz constant} of the function $f$ \cite{o2006metric}. A corollary that arises from Lipschitz Continuity is that if the gradient of a function is bounded:
\begin{equation}
    \| \nabla_{\rvx}f(\rvx) \| \leq K,
    \label{eq: lipschitz-gradient}
\end{equation}
then this function $f$ is Lipschitz continuous. However, it is worth noting that the converse is not true.
\end{definition}

\subsection{Reinforcement Learning}
In this work, our controllers are trained through reinforcement learning, in which an agent interacts with the environment according to a policy $\pi$ to maximize an objective function \cite{sutton2018reinforcement}. At each timestep $t$, the agent observes the state $\rvs_t$ of the environment, and takes an action $\rva_t$ according to the policy $\pi(\rva_t~|~\rvs_t)$. This action then leads to a new state according to the dynamics of the environment $p(\rvs_{t+1}~|~\rvs_t,\rva_t)$. The agent receives a reward $r_t=r(\rvs_{t+1},\rvs_t,\rva_t)$ at each step. The agent's goal is to maximize its expected return:
\begin{equation}
J(\pi)=\mathbb{E}_{p(\tau \mid \pi)}\left[\sum_{t=0}^{T-1} \gamma^t r_t\right],
\label{eq:RL-obj}
\end{equation}
where $p(\tau|\pi)$ represents the likelihood of the trajectory $\tau$, $T$ denotes the time horizon, and $\gamma$ is the discount factor.

\section{Lipschitz-Constrained Policies}
In this section, we introduce Lipschitz-Constrained Policies (\ours), a method for training policies to produce smooth behaviors by incorporating a Lipschitz constraint during training. We begin with a simple experiment to illustrate the motivation behind our method. This is then followed by a detailed description of our proposed method.


\begin{figure}[t]
        \centering
        \includegraphics[width=0.98\columnwidth]{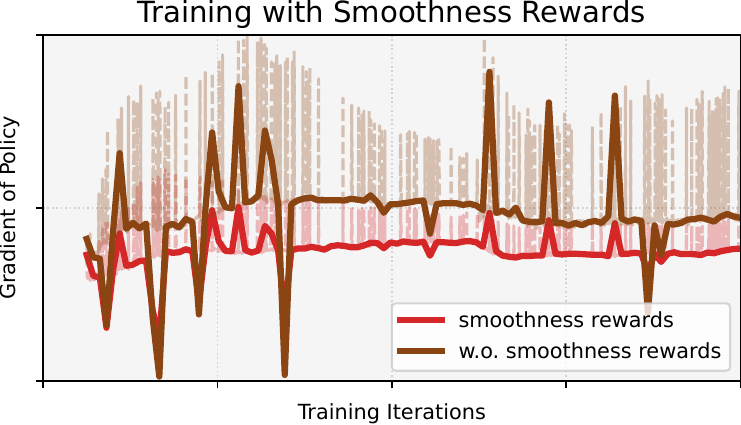}
        \caption{Gradient of policies trained with and without smoothness rewards. Policies with smoother behaviors also exhibit smaller gradient magnitudes.}
        \label{fig:intuitive-exp-2}
\vspace{-0.25cm}
\end{figure}


\subsection{Motivating Example}
We will first illustrate the motivation of \ours with a simple experiment. We know that RL-based policies are prone to producing jittery behaviors, and the most common method for mitigating these behaviors is to incorporate smoothness rewards during training.
The smoothness of a function is typically evaluated using the first derivative. Therefore, we compare the \(\ell^2\)-norm of the gradient of policies trained with and without smoothness rewards. Although no specific technique is explicitly applied to regularize the policies' gradient, the gradient trained with smoothness rewards is significantly smaller than that of a policy trained without smoothness rewards, as illustrated in \Figref{fig:intuitive-exp-2}. This fact inspires our proposed method, which explicitly regularizes the gradient of the policy. 
We show that this simple method leads to smooth behaviors, which can then facilitate successful transfer to the real world.

\subsection{Lipschitz Constraint as a Differentiable Objective}
While smoothness rewards can mitigate jittery behaviors, these reward functions can be complex to design, with a large number of hyperparameters that require tuning. Furthermore, these smoothness rewards are non-differentiable since they are implemented as part of the underlying environment. Therefore, they often need to be optimized through sampling-based methods, such as policy gradients. This work proposes a simple and differentiable smoothness objective for policy optimization based on Lipschitz continuity.

\Eqref{eq: lipschitz-gradient} stipulates that any function with bounded gradients is Lipschitz continuous. Therefore, we can formulate a constrained policy optimization problem that enforces Lipschitz continuity through a gradient constraint:
\begin{equation}
\begin{aligned}
\max_{\pi} \quad & J(\pi)\\
\textrm{s.t.} \quad & \mathop{\mathrm{max}}_{\rvs, \rva} \left[\| \nabla_{\rvs}\log\pi(\rva|\rvs) \|^2 \right]  \leq K^2 \\
\end{aligned}
\label{eq:constraint-lipschitz}
\end{equation}
where $K$ is a constant and $J(\pi)$ is the RL objective defined in \Eqref{eq:RL-obj}. Since calculating the maximum gradient norm across all states is intractable, we approximate this constraint with an expectation over samples collected from policy rollouts, following the heuristic from \textit{Schulman et al.}  \cite{TRPOSchulman2015}:
\begin{equation}
\begin{aligned}
\max_{\pi} \quad & J(\pi)\\
\textrm{s.t.} \quad & \mathbb{E}_{\rvs, \rva\sim\mathcal{D}}\left[ \| \nabla_{\rvs} \log\pi(\rva|\rvs) \|^2 \right] \leq K^2, \\
\end{aligned}
\label{eq:constraint-lipschitz-sample}
\end{equation}
where $\mathcal{D}$ is a dataset consisting of state-action pairs $(\rvs_t, \rva_t)$ collected from the policy. Next, to facilitate optimization with gradient-based methods, the constraint can be reformulated into a penalty by introducing a Lagrange multiplier $\lambda$:
\begin{equation}
    \min_{\lambda \geq 0} \max_{\pi} \quad J(\pi) - \lambda \left(\mathbb{E}_{\rvs, \rva\sim\mathcal{D}}\left[ \| \nabla_{\rvs} \log\pi(\rva|\rvs) \|^2 \right] - K^2\right).
\end{equation}
To further simplify the objective, we set $\lambda_\mathrm{gp}$ as a manually specified coefficient, and since $K$ is a constant, this leads to a simple differentiable gradient penalty (GP) on the policy:
\begin{equation}
    \max_{\pi} \quad J(\pi) - \lambda_\mathrm{gp} \mathbb{E}_{\rvs, \rva\sim\mathcal{D}}\left[ \| \nabla_{\rvs} \log\pi(\rva|\rvs) \|^2 \right].
    \label{eq:lcp-obj}
\end{equation}
This gradient penalty can be easily implemented in any reinforcement learning framework, requiring only a few lines of code.
The gradient penalty provides a simple and differentiable alternative to smoothness rewards or low-pass filters, which are not differentiable with respect to the policy parameters. Our experiments show that \ours provides an effective alternative to non-differentiable smoothing techniques and can be directly used to train robust locomotion controllers for a diverse cast of robots.

\section{Training Setup}
To evaluate the effectiveness of our method, we apply LCP to train policies for a variety of humanoid robots, where the task is for the robots to walk while following steering commands.



\paragraph{Observations} The input observations to the policy $\rvo_t = [\boldsymbol{\phi}_t, \rvc_t, \rvs_{t}^{\text{robot}}, \rva_{t-1}]$ consists of a gait phase variable $\boldsymbol{\phi}_t\in\mathbb{R}^2$ (a periodic clock signal represented by its sine and cosine components),
command $\rvc_t$, measured joint positions and velocities $\rvs_{t}^{\text{robot}}$, and the previous output action of the policy $\rva_{t-1}$. To enable robust sim-to-real transfer, the policy also takes privileged information $\mathbf{e}_t$ as input, which consists of the base mass, center of mass, motor strengths, and root linear velocity. Observations $\textbf{o}_t$ are normalized with a running mean and standard deviation before being passed as input to the policy.

\paragraph{Commands} The command input to the policy $\textbf{c}_t = [\rv_{\text{x}}^{\text{cmd}}, \rv_{\text{y}}^{\text{cmd}}, \rv_{\text{yaw}}^{\text{cmd}}]$ consists of the desired linear velocities along x-axis $\rv_{\text{x}}^{\text{cmd}}\in [0\text{m/s}, 0.8\text{m/s}]$ and y-axis $\rv_{\text{y}}^{\text{cmd}}\in [-0.4\text{m/s}, 0.4\text{m/s}]$, and the desired yaw velocity $\rv_{\text{yaw}}^{\text{cmd}}\in [-0.6\text{rad/s}, 0.6\text{rad/s}]$, both are in the robot frame. During training, commands are randomly sampled from their respective ranges every $150$ timestep or when the environment is reset.

\paragraph{Actions} The policy's output actions specify target joint rotations for all joints in the robot's body, which are then converted to torque commands by PD controllers with manually specified PD gains.

\paragraph{Training} All policies are modeled using neural networks and trained using the PPO algorithm \cite{schulman2017proximal}. The policies are trained solely in simulation with domain randomization and then deployed directly on the real robots \cite{Sim2Real2018}. Sim-to-real transfer is performed using Regularized Online Adaptation (ROA) \cite{fu2023deep, liu2024visual}.


\input{sec/4_exp}

\section{Conclusion}
In this work, we present Lipschitz Constrained Policies (\ours), a simple and general method for training controllers to produce smooth behaviors amenable to sim-to-real transfer. \ours approximates a Lipschitz constraint on the policy, implemented in the form of a differentiable gradient penalty applied during training. Through extensive simulation and real-world experiments, we show the effectiveness of \ours in training locomotion controllers for a wide range of real humanoid robots.
While \ours has demonstrated its effectiveness in real-world locomotion experiments, our results are still limited to basic walking behaviors. Evaluating \ours on more dynamic skills, such as running and jumping, would help further validate this method's generality.

\section*{Acknowledgement}
We extend our gratitude to Fourier Intelligence for their hardware support, Jiaze Cai, Yiyang Shao, Junfeng Long, Haoru Xue, and Renzhi Tao for their assistance with real-world experiments, and Mintae Kim for his valuable advice on the mathematical formulas. This work was supported in part by an NSERC Discovery Grant (DGECR-2023-00280).


\balance
{
\bibliographystyle{IEEEtran}
\bibliography{main}
}


\input{sec/X_appendix}

\end{document}

%% file: math_commands.tex
\def\Tabref#1{TABLE~\ref{#1}}

\def\Figref#1{Fig.~\ref{#1}}

\def\eqref#1{equation~\ref{#1}}
\def\Eqref#1{Equation~\ref{#1}}

\def\1{\bm{1}}

\def\rv{{\textnormal{v}}}

\def\rva{{\mathbf{a}}}

\def\rvc{{\mathbf{c}}}

\def\rve{{\mathbf{e}}}

\def\rvo{{\mathbf{o}}}

\def\rvs{{\mathbf{s}}}

\def\rvx{{\mathbf{x}}}

\def\rvz{{\mathbf{z}}}

\DeclareMathAlphabet{\mathsfit}{\encodingdefault}{\sfdefault}{m}{sl}
\SetMathAlphabet{\mathsfit}{bold}{\encodingdefault}{\sfdefault}{bx}{n}



%% file: sec/4_exp.tex
\begin{figure*}[ht]
\centering
\includegraphics[width=\textwidth]{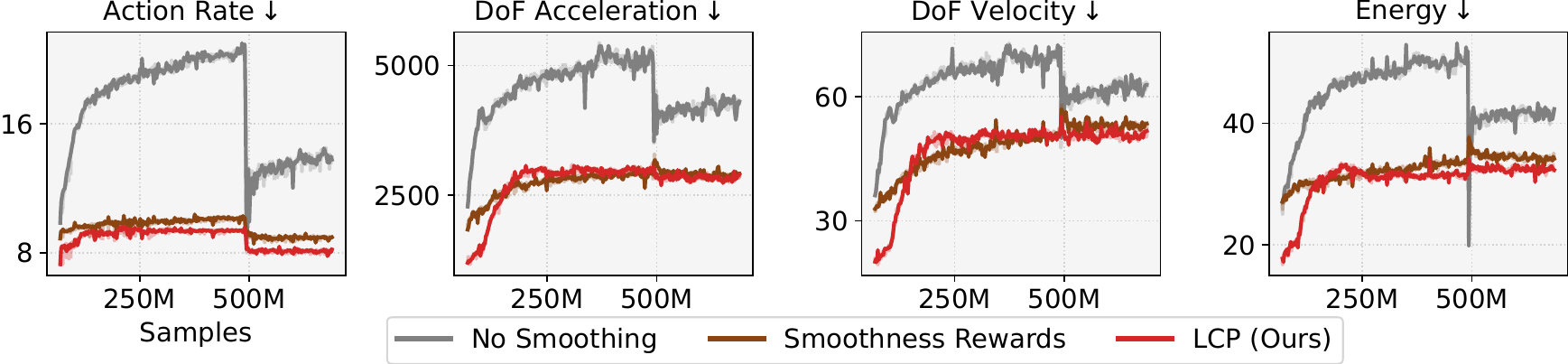}
\caption{Smoothness metrics recorded over the course of training. \ours produces smooth behaviors that are comparable to policies that are trained with explicit smoothness rewards.}
\label{fig:substitute-exp}
\end{figure*}

\begin{figure*}[t]
    \centering
    \begin{minipage}[t]{0.45\textwidth}
    \centering
    \includegraphics[width=0.95\textwidth]{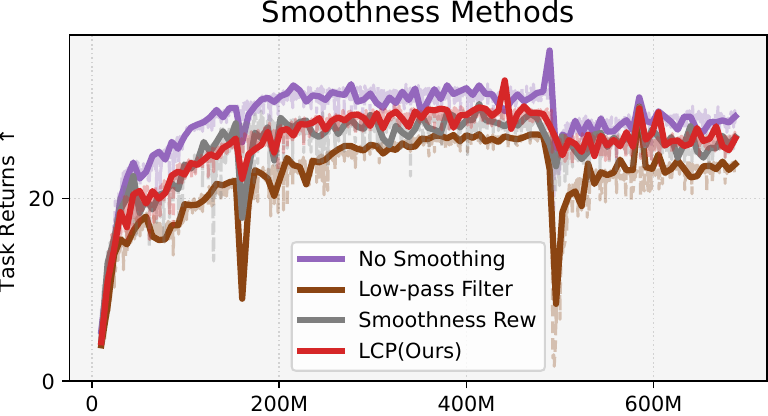}
    \caption{Task returns of different smoothing methods. \ours provides an effective alternative to other techniques.}
    \label{fig:sub-task-return-smooth-methods}
    \end{minipage}
    \hspace{0.2cm}
    \begin{minipage}[t]{0.45\textwidth}
    \centering
    \includegraphics[width=0.95\textwidth]{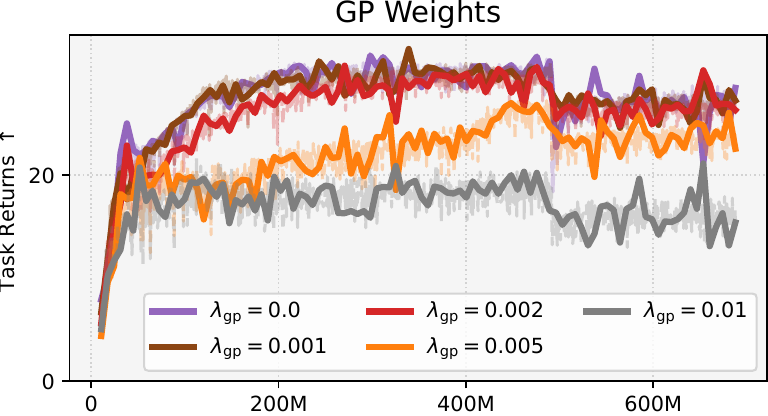}
    \caption{Task returns of \ours with different $\lambda_{\text{gp}}$. Excessively large $\lambda_{\text{gp}}$ may hinder policy learning.}
    \label{fig:sub-task-return-gp-weights}
    \end{minipage}
    \vspace{-0.5cm}
\end{figure*}

\section{Experiments}

LCP's effectiveness is evaluated on a set of diverse humanoid robots to show its generalization ability. We conduct an extensive suite of simulation and real-world experiments, comparing LCP to commonly used smoothing techniques from prior systems.

\subsection{Robot Platforms}
We evaluate our framework on three real-world robots: the human-sized Fourier GR1T1, Fourier GR1T2, Unitree H1, and the smaller Berkeley Humanoid. We will first provide an overview of each robot's body structure. Then, our experiments show that \ours is a general smoothing technique that can be applied widely to several distinct robots.

\paragraph{Fourier GR1T1 \& Fourier GR1T2}
The Fourier GR1T1 and Fourier GR1T2 have the same mechanical structure. They both comprise $21$ joints, with $12$ joints in the lower body and $9$ in the upper body. Notice that the torque limit for the \textit{ankle roll} joint is minimal; we treat this as a passive joint during training and deploying. This means we control $19$ joints of GR1 in total.

\paragraph{Unitree H1}
The Unitree H1 has $19$ joints, with $10$ joints in the lower body, $9$ in the upper body, and $1$ ankle joint per leg. All joints are actively controlled.

\paragraph{Berkeley Humanoid}
Berkeley Humanoid is a small robot with a height of $0.85$m \cite{liao2024berkeley}.
It has $12$ degrees of freedom, with $6$ joints in each leg and $2$ joints in each ankle. 

\subsection{Results}

\input{tables/sim_berkeley_ablations}

To evaluate the effectiveness of \ours, we compare our method to the following baselines:
\begin{itemize}
    \item \textbf{No smoothing:} No smoothing techniques are applied during training. This baseline demonstrates the necessity of smoothing techniques for sim-to-real transfer;
    \item \textbf{Smoothness rewards:} Smoothness rewards are the most commonly used smoothing method, where additional reward terms are incorporated into the reward function to encourage smooth behaviors. These reward functions are not directly differentiable.
    \item \textbf{Low-pass Filters:} Low-pass filters are commonly used for action smoothing, where a filter is applied to the policy's output actions before the action is applied to the environment. Low-pass filters are also not easily differentiable for policy training.
\end{itemize}
To evaluate the effectiveness of various smoothing techniques, we record a suite of smoothness metrics, including mean DoF velocities (rad/s), mean energy (N$\cdot$rad/s), action rate (rad/s), robot base acceleration (m/s$^2$), action jitter (rad/s$^3$), and DoF position jitter (rad/s$^3$). Action rate is the first derivative of output actions over time. The jitter metrics represent the third derivative of their respective quantities \cite{flash1985coordination}. 
We also record the mean task return, which is calculated using the linear and angular velocity tracking rewards.

\input{tables/sim2sim_performance}

\paragraph{Is \ours effective for producing smooth behaviors?} We train  policies with \ours using a GP coefficient of $\lambda_{\text{gp}} = 0.002$. We track various smoothness metrics throughout the training process, including energy consumption, degrees-of-freedom (DoF) velocities, DoF accelerations, and action rates. We then compare these metrics against policies trained with and without smoothness rewards. The results are recorded in \Figref{fig:substitute-exp}. While \ours is not trained with reward functions that directly minimize these smoothness metrics, \ours nonetheless produces smooth behaviors that are similar to policies trained with smoothness rewards. This demonstrates that \ours can be an effective substitute for traditional smoothness rewards.

\begin{figure}[ht]
\centering
\includegraphics[width=\columnwidth]{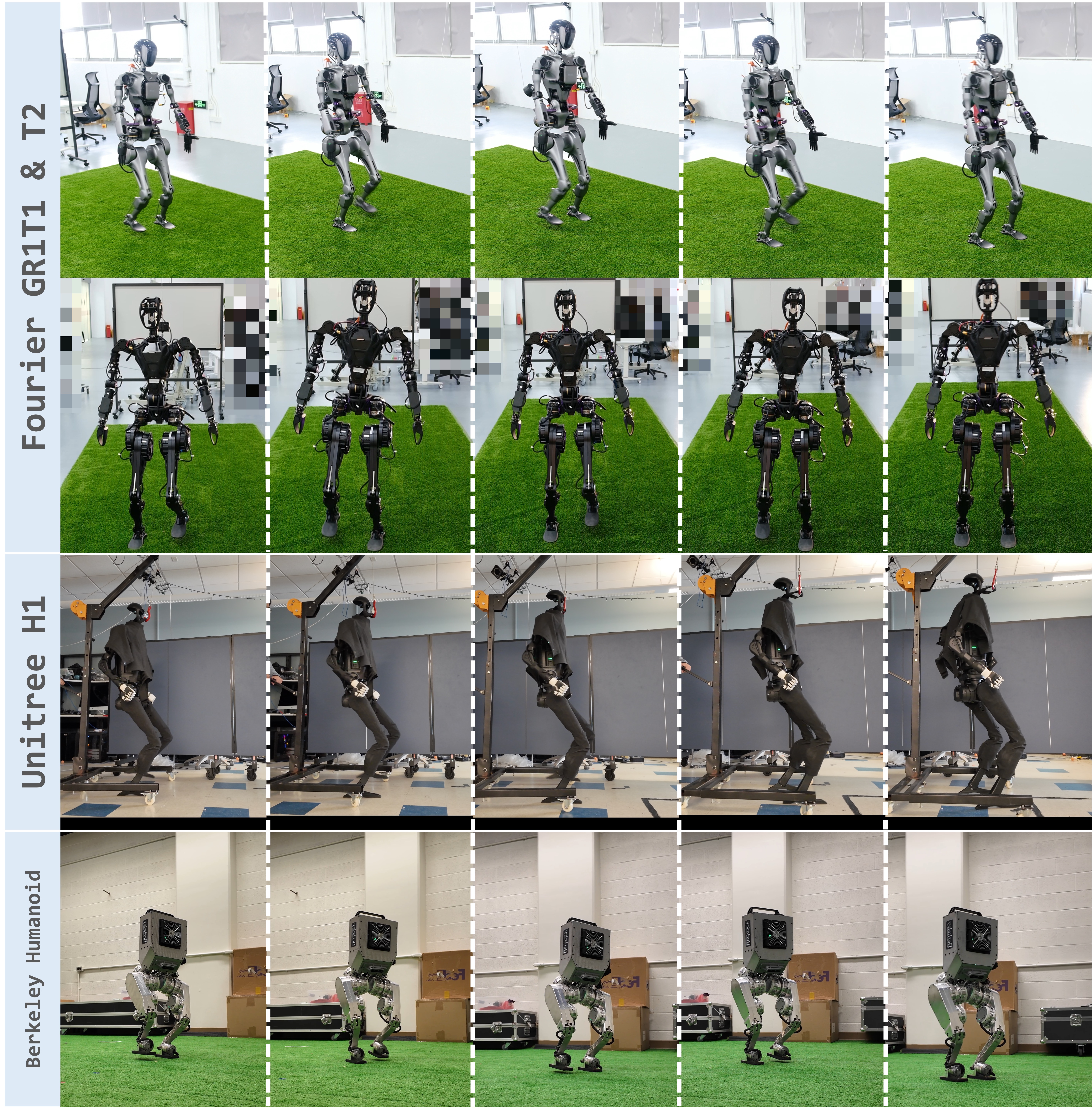}
\caption{Real-world deployment. \ours is able to train effective locomotion policies on a wide range of robots, which can be directly transferred to the real world.}
\label{fig:robot-walking}
\vspace{-0.5cm}
\end{figure}

\paragraph{How does \ours affect task performance?} In \Tabref{tab:ablations-berkeley}(a) and \Figref{fig:sub-task-return-smooth-methods}, we compare the task performance of \ours with policies trained with other smoothing methods. \ours achieves similar task performance compared to policies trained solely with smoothness rewards. Policies trained with low-pass filters tend to exhibit lower task returns, which may be due to the damping introduced by low-pass filters that can in turn impair exploration. Policies trained without smoothing techniques tend to achieve the highest task returns but exhibit highly jittery behaviors unsuitable for real-world deployment.

\paragraph{What is the effect of the GP coefficient $\lambda_{\text{gp}}$?} \Tabref{tab:ablations-berkeley}(b) shows the performance of \ours with different GP coefficients $\lambda_{\text{gp}}$. Incorporating a gradient penalty leads to significantly smoother behaviors. However, with small coefficients (e.g., $\lambda_{\text{gp}}=0.001$), the policy can develop jittery behaviors that are dangerous to deploy in the real world. However, excessively larger coefficients (e.g., $\lambda_{\text{gp}}=0.01$) lead to a substantial decline in task return due to overly smooth and sluggish behaviors. As shown in \Figref{fig:sub-task-return-gp-weights}, large values of $\lambda_{\text{gp}}$ may also lead to slower learning speeds. Our experiments suggest that $\lambda_{\text{gp}}=0.002$ strikes an effective balance between policy smoothness and task performance. As with other smoothing techniques, some care is required to tune the GP coefficient for a better performance.

\paragraph{Which components of the observation should GP be applied to?} Since the policies are trained using the ROA for sim-to-real transfer, the policy's input consists of the current observation and a history of past observations. \Tabref{tab:ablations-berkeley}(c) compares the performance of \ours when the gradient penalty is applied to the whole input observation or only to the current observation. We find that applying GP on the whole observation achieves the best performance. Regularizing the policy only with respect to the current observation can still lead to non-smooth behaviors due to changes in the history.

\paragraph{Sim-to-Sim Transfer} Before deploying models we test our models in a different simulator Mujoco \cite{todorov2012mujoco}. As shown in \Tabref{tab:sim2sim_performance_table}, we observe a slight decrease in task return compared to IsaacGym for full-sized robots such as Fourier GR1 and Unitree H1, suggesting that the domain gap is more significant for larger robots.
The overall results show that \ours performs well in sim-to-sim transfer, providing confidence for subsequent real-world deployments.

\input{tables/real_robot_performance}
\subsection{Real World Deployment}
We deploy \ours models trained with the same reward functions and $\lambda_{\text{gp}}=0.002$ on four distinct robots. As shown in \Figref{fig:teaser}, \ours effectively enables different robots to walk in the real world. \Figref{fig:robot-walking} shows snapshots of the robots' behaviors over the course of one gait cycle.

\noindent\textbf{Terrains} To evaluate the robustness of the learned policies, we apply the policies in the real world to walk on three types of terrain: smooth, soft, and rough plane. We measure the jitter metrics to evaluate \ours's performance, as shown in \Tabref{tab:realrobot_performance_table}. The behaviors of our policies remain smooth in the presence of variations in the terrain and is able to effective traverse the different surfaces.

\noindent\textbf{External Forces} To further test the robustness of our policies, we apply external forces to the robot in the real world \Figref{fig:teaser}. The recovery behaviors of our models are shown in the supplementary video. The \ours models can robustly recover from unexpected external perturbations.


%% file: tables/sim_berkeley_ablations.tex
\begin{table*}[!ht]
\caption{\textbf{Ablation Studies.} All policies are trained with three random seeds and tested in $1000$ environments for $500$ steps, corresponding to $10$ seconds clock time.}
\label{tab:ablations-berkeley}
\centering
\renewcommand{\arraystretch}{0.8}
\begin{tabular}{lcccccc}
\toprule
\textbf{Method} & \textbf{Action Jitter} $\downarrow$ & \textbf{DoF Pos Jitter} $\downarrow$ & \textbf{DoF Velocity} $\downarrow$ & \textbf{Energy} $\downarrow$ & \textbf{Base Acc} $\downarrow$ & \textbf{Task Return} $\uparrow$ \\
\cmidrule(r){1-1}\cmidrule(r){2-7}
\rowcolor{lightgray} 
\multicolumn{1}{l}{\textbf{(a) Ablation on Smooth Methods}} &&&&&&\\
\cmidrule(r){1-1} \cmidrule(r){2-7}
\textbf{\ours (ours)} & $\mathbf{3.21\pm 0.11}$ & $\mathbf{0.17\pm 0.01}$ & $\mathbf{10.65\pm 0.37}$ & $\mathbf{24.57\pm 1.17}$ & $\mathbf{0.06 \pm 0.002}$ & $26.03\pm 1.51$ \\
\textbf{Smoothness Reward} & $5.74\pm 0.08$ & $0.19\pm 0.002$ & $11.35 \pm 0.51$ & $25.92\pm 0.84$  & $\mathbf{0.06\pm 0.002}$ & $26.56 \pm 0.26$\\
\textbf{Low-pass Filter} & $7.86\pm3.00$ & $0.23\pm0.04$ & $11.72\pm0.14$ & $32.83\pm5.50$  & $\mathbf{0.06\pm0.002}$ & $24.98\pm1.29$\\
\textbf{No Smoothness} & $42.19 \pm 4.72$ & $0.41 \pm 0.08$ & $12.92\pm 0.99$ & $42.68\pm 10.27$  & $0.09\pm 0.01$ & $\mathbf{28.87\pm 0.85}$\\
\cmidrule(r){1-1} \cmidrule(r){2-7}
\rowcolor{lightgray} 
\multicolumn{1}{l}{\textbf{(b) Ablation on GP Weights ($\lambda_{\text{gp}}$)}} &&&&&&\\
\cmidrule(r){1-1} \cmidrule(r){2-7}
\textbf{\ours w.} $\lambda_{\text{gp}}=0.0$ & $42.19 \pm 4.72$ & $0.41 \pm 0.08$ & $12.92\pm 0.99$ & $42.68\pm 10.27$  & $0.09\pm 0.01$ & $\mathbf{28.87\pm 0.85}$\\
\textbf{\ours w.} $\lambda_{\text{gp}}=0.001$ & $3.69 \pm 0.31$ & $0.21\pm 0.05$ & $11.44\pm 1.18$ & $27.09\pm 4.44$ & $0.06 \pm 0.01$& $26.32\pm 1.20$\\
\textbf{\ours w.} $\lambda_{\text{gp}}=0.002$ (ours) & $3.21\pm 0.11$ & $0.17\pm 0.01$ & $10.65\pm 0.37$ & $24.57\pm 1.17$ & $0.06 \pm 0.002$ & $26.03\pm 1.51$\\
\textbf{\ours w.} $\lambda_{\text{gp}}=0.005$ & $2.10 \pm 0.05$ & $ 0.15\pm 0.01$ & $10.44\pm 0.70$ & $26.24\pm 3.50$ & $0.05\pm 0.002$ & $23.92\pm 2.05$\\
\textbf{\ours w.} $\lambda_{\text{gp}}=0.01$ & $\mathbf{0.17 \pm 0.01}$ & $ \mathbf{0.07\pm 0.00}$ & $\mathbf{2.75\pm 0.12}$ & $\mathbf{5.89\pm 0.28}$ & $\mathbf{0.007\pm 0.00}$ & $16.11\pm 2.76$\\
\cmidrule(r){1-1} \cmidrule(r){2-7}
\rowcolor{lightgray} 
\multicolumn{1}{l}{\textbf{(c) Ablation on GP Inputs}} &&&&&&\\
\cmidrule(r){1-1} \cmidrule(r){2-7}
\textbf{\ours w. GP on whole obs (ours)}& $\mathbf{3.21\pm 0.11}$ & $\mathbf{0.17\pm 0.01}$ & $\mathbf{10.65\pm 0.37}$ & $\mathbf{24.57\pm 1.17}$ & $\mathbf{0.06 \pm 0.002}$ & $\mathbf{26.03\pm 1.51}$ \\
\textbf{\ours w. GP on current obs}& $7.16\pm 0.60$ & $0.35\pm0.03$ & $13.70\pm1.50$ & $35.18\pm 4.84$ & $0.09\pm0.005$ & $25.44\pm3.73$\\
\bottomrule 
\end{tabular}
\end{table*}

%% file: tables/sim2sim_performance.tex
\begin{table*}[!ht]
\caption{Sim-to-sim perfomance when transferring policies trained in IsaacGym to Mujoco. All policies are trained with three random seeds and tested for $3$ trials with $500$ steps, corresponding to $10$ seconds per trial.}
\label{tab:sim2sim_performance_table}
\centering
\renewcommand{\arraystretch}{0.8}
\begin{tabular}{lcccccc}
\toprule
~ & \textbf{Action Jitter} $\downarrow$ & \textbf{DoF Pos Jitter} $\downarrow$ & \textbf{DoF Velocity} $\downarrow$ & \textbf{Energy} $\downarrow$ & \textbf{Base Acc} $\downarrow$ & \textbf{Task Return} $\uparrow$  \\
\midrule
\textbf{Fourier GR1} & $1.47\pm0.43$ & $0.34\pm0.07$ & $9.54\pm1.53$ & $36.38\pm2.97$ & $0.08\pm 0.004$ & $24.33\pm1.25$  \\
\textbf{Unitree H1} & $0.44\pm0.03$ & $0.10\pm0.007$ & $9.12\pm0.38$ & $76.22\pm 5.81$ & $0.04\pm 0.005$ & $21.74\pm 1.40$  \\
\textbf{Berkeley Humanoid} & $1.77\pm0.32$ & $0.12\pm0.01$ & $7.92\pm0.21$ & $19.99\pm0.36$ & $0.06\pm0.00$ & $26.50\pm 0.57$  \\
\bottomrule 
\end{tabular}
\end{table*}

%% file: tables/real_robot_performance.tex
\begin{table}[t]
\caption{Performance during real-world deployment. Performance for each method is calculated across 3 models from different training runs. Each model is executed for $10$ seconds. Standard deviation is recorded for each test.}
\label{tab:realrobot_performance_table}
\centering
\renewcommand{\arraystretch}{0.8}
\begin{adjustbox}{width=1\columnwidth, center}
\begin{tabular}{lccc}
\toprule
\textbf{Robot} & \textbf{Action Jitter} $\downarrow$ & \textbf{DoF Pos Jitter} $\downarrow$ & \textbf{DoF Velocity} $\downarrow$ \\
\cmidrule(r){1-1}\cmidrule(r){2-4}
\rowcolor{lightgray}
\multicolumn{1}{l}{\textbf{(a) Smooth Plane}} &&&\\
\cmidrule(r){1-1} \cmidrule(r){2-4}
\textbf{Fourier GR1} & $1.12\pm 0.16$ & $0.28\pm 0.13$ & $10.82\pm 1.58$ \\
\textbf{Unitree H1} & $1.11\pm 0.07$ & $0.14\pm 0.01$ & $10.95\pm 0.53$ \\
\textbf{Berkeley Humanoid} & {$1.56 \pm 0.10$} & {$0.10 \pm 0.01$} & {$4.99 \pm 0.60$} \\
\cmidrule(r){1-1} \cmidrule(r){2-4}
\rowcolor{lightgray}
\multicolumn{1}{l}{\textbf{(b) Soft Plane}} &&&\\
\cmidrule(r){1-1} \cmidrule(r){2-4}
\textbf{Fourier GR1} & $1.18\pm 0.17$ & $0.24\pm 0.09$ & $10.45 \pm 1.42$ \\
\textbf{Unitree H1} & $1.18\pm 0.09$ & $0.15\pm 0.01$ & $11.80\pm 0.57$ \\
\textbf{Berkeley Humanoid} & {$1.66 \pm 0.03$} & {$0.12 \pm 0.01$} & {$6.78 \pm 1.57$}\\
\cmidrule(r){1-1} \cmidrule(r){2-4}
\rowcolor{lightgray} 
\multicolumn{1}{l}{\textbf{(c) Rough Plane}} &&&\\
\cmidrule(r){1-1} \cmidrule(r){2-4}
\textbf{Fourier GR1} & $1.18\pm 0.22$ & $0.26\pm 0.11$ & $11.61\pm 1.64$ \\
\textbf{Unitree H1} & $1.20\pm 0.09$ & $0.14\pm 0.01$ & $11.68\pm 0.84$ \\
\textbf{Berkeley Humanoid} & {$1.63 \pm 0.11$} & {$0.11 \pm 0.01$} & {$5.02 \pm 0.48$} \\
\bottomrule 
\end{tabular}
\end{adjustbox}
\vspace{-0.4cm}
\end{table}

%% file: sec/X_appendix.tex
\clearpage
\newpage
\onecolumn
\begin{appendix}

\subsection{Regularized Online Adaptation}
During training with ROA, an encoder $\mu$ embeds the privileged information $\rve$ to an environment extrinsic latent vector $\rvz^{\mu}$. An adaptation module $\phi$ then estimates this embedding $\rvz^{\mu}$ based only on the recent history of proprioceptive observations. Along with the Lipschitz constraint specified in \Eqref{eq:lcp-obj}, we can formulate the training loss as follows:
\begin{equation}
\begin{split}
    L(\theta_{\pi}, \theta_{\mu}, \theta_{\phi}) &= -L^{PPO}(\theta_{\pi}, \theta_{\mu}) + \lambda \left\| \rvz^{\mu} - \text{sg}\left[ \rvz^{\phi} \right] \right\| \\ &+ \left\| \text{sg}\left[\rvz^{\mu}\right] - \rvz^{\phi} \right\| + \lambda_{\text{gp}} L_{\text{gp}}(\pi),
    \label{eq:overall-loss}
\end{split}
\end{equation}
where $\text{sg}[\cdot]$ is the stop gradient operator, $L_{\text{gp}}(\pi)$ is defined as:
\begin{equation}
    L_{\text{gp}}(\pi) = \mathbb{E}_{\rvs, \rva\sim\mathcal{D}}\left[ \| \nabla_{\rvs} \log\pi(\rva|\rvs) \|^2 \right],
\end{equation}
we set $\lambda_{\text{gp}} = 0.002, \lambda=0.1$ during the training process.

\subsection{Reward Curriculum}
We have revised the cumulative discounted reward formulation to incorporate a curriculum. Specifically, the updated expression is
$\mathbb{E}\left[\sum_{t=1}^T \gamma^{t-1} \sum_{i} s_{t,i} r_{t,i}\right]$, 
where $r_{t,i}$ denotes different reward components at time $t$, and $s_{t,i}$ is a scaling factor defined as follows: 
$$
s_{t,i} = \begin{cases} 
s_{\text{current}} & \text{if } r_{t,i} < 0 \\
1 & \text{if } r_{t,i} \geq 0 
\end{cases},
$$
where $s_{\text{current}}$ is dynamically adjusted, starting from an initial value of $0.8$. If the average episode length is below $50$, the scaling factor is reduced by multiplying by $0.9999$. Conversely, if the episode length exceeds $400$, it is increased by multiplying by $1.0001$. The scaling factor is capped at an upper limit of $2.0$. The intuition behind this is using smaller regularization rewards at the start for exploration and larger regularization later for desired behaviors.

\subsection{Training Details}
We use IsaacGym as the simulator. The number of parallel environments is $4096$. Our reward functions consist of \textit{gait style rewards}, \textit{velocity tracking rewards}, and \textit{regularization rewards}, where gait style rewards and velocity tracking rewards are usually task-specific. Here we provide the details of regularization reward functions in \Tabref{tab:training-details}. For specific implementations of all the reward functions, please refer to our \href{https://github.com/zixuan417/smooth-humanoid-locomotion}{\color{deepblue}{codebase}}.

\begin{table*}[ht]
\centering
\caption{Terms and weights of regularization rewards.}
\begin{tabular}{ll}
\toprule
\textbf{Name} & \textbf{Weight} \\
\midrule
Angular velocity $xy$ penalty & \( 0.2 \) \\
Joint torques & \( 6e-7 \) \\
Collisions & \( 10 \) \\
Linear velocity $z$ & $-1.5$ \\
Contact force penalty & $-0.002$ \\
Feet Stumble & $-1.25$ \\
Dof position limit & $-10$ \\
Base Orientation & $-1.0$ \\
\bottomrule
\end{tabular}
\label{tab:training-details}
\end{table*}

\end{appendix}